\def\year{2022}\relax
\documentclass[letterpaper]{article}
\usepackage{aaai22} 
\usepackage{times} 
\usepackage{helvet} 
\usepackage{courier} 
\usepackage[hyphens]{url} 
\usepackage{graphicx} 
\usepackage{graphicx} 
\usepackage{natbib} 
\usepackage{caption} 
\DeclareCaptionStyle{ruled}%
{labelfont=normalfont,labelsep=colon,strut=off}
\usepackage{eqnarray}

\usepackage{subcaption}
\usepackage{xcolor}
\usepackage[linesnumbered,ruled,vlined]{algorithm2e}
\usepackage[ansinew]{inputenc}

\usepackage{amssymb}
\usepackage{amsmath}

\usepackage{cite}

\SetCommentSty{mycommfont}

\usepackage{amsfonts}

\newtheorem{theorem}{Theorem}

\begin{document}
%

\author{Nicolas Cavrel, Damien Pellier, Humbert Fiorino,\\
{\normalfont Univ. Grenoble Alpes - LIG}\\
{\normalfont Grenoble, France}\\
{\normalfont \{nicolas.cavrel, damien.pellier, humber.fiorino\}@univ-grenoble-alpes.fr}}

\title{An Efficient HTN to STRIPS Encoding for Concurrent Plans}

\maketitle
\begin{abstract}
\begin{quote}

The Hierarchical Task Network (HTN) formalism is used to express a wide variety of planning problems in terms of decompositions of tasks into subtaks. Many techniques have been proposed to solve such hierarchical planning problems. A particular technique is to encode hierarchical planning problems as classical STRIPS planning problems. One advantage of this technique is to benefit directly from the constant improvements made by STRIPS planners. However, there are still few effective and expressive encodings. In this paper, we present a new HTN to STRIPS encoding allowing to generate {\it concurrent} plans. We show experimentally that this encoding outperforms previous approaches on hierarchical IPC benchmarks.
\end{quote}
\end{abstract}

\section{Introduction}

The Hierarchical Task Network (HTN) formalism \cite{complexityHTN} is used to express a wide variety of planning problems in terms of decompositions of tasks into subtaks. HTN planning is used in many applications as, for instance, in task allocation for robot fleets \cite{htnTaskAuction}, video games \cite{htnVideoGames} or industrial contexts such as software deployment \cite{htnIndus}. One possible explanation is that HTN formalism usually fits better for real-world applications and domain experts' mindset: a HTN planning problem is expressed as a set of tasks to achieve rather than an objective state to reach, and the "processes" achieving these tasks as \textit{methods}, that is to say task decompositions into "simplier" subtasks. Despite the success of hierarchical planning and the recent revival of this planning technique \cite{BercherAH19}, there is comparatively less work on hierarchical planning than in classical STRIPS planning \cite{FIKES1971189}. The work of the planning community has been more focused on the development of techniques and heuristics for STRIPS planning, e.g., \cite{ffHeuristic, heuristic1, heuristic2}.

Many techniques are used to solve hierarchical planning problems. Some ad-hoc HTN solvers have been implemented \cite{poclhtn, Erol1996HierarchicalTN, shop2}. Another approach consists in encoding HTN problems into SAT problems \cite{htn2sat} or into constraint programming problems \cite{cpt}.

One particular technique of encoding is to translate hierarchical planning problems as STRIPS problems. Encoding techniques benefit directly from the constant improvements of STRIPS planners. To our best knowledge, two HTN to STRIPS encodings have been published so far \cite{Alford2016BoundTP, translation1} (very recent work have been published concurrently to this paper, which will not be studied here \cite{Behnke2022Translation}). However, they have some limitations on the type of problems they can address and only produce {\it sequential} plans. For instance, one of the best current encodings is  \cite{Alford2016BoundTP}. It translates any HTN problem into STRIPS, making it solvable by any STRIPS planner. However, this encoding has three downsides:
\begin{enumerate}
    \item It depends on a \textit{progression bound}, which is an integer parameter bounding the size of the task network, meaning that the maximum size of the task network has to be estimated before encoding the HTN problem into STRIPS.
    \item The resulting solution plans are sequential, a single action being performed at each time. However, many real-world applications are intrinsically distributed, and need \textit{concurrent} actions at each time.
    \item The encoding generates an high number of actions in the translated STRIPS problem. This makes the actual grounding of the problem difficult. This is particularly true when the makespan increases, as the number of translated actions greatly expands with it.
\end{enumerate}

The contribution of the paper is twofold: (1) we introduce a search procedure called CPFD (Concurrent Partial Forward Decomposition) to generate concurrent plans, and (2) propose an {\bf encoding} of this search procedure into STRIPS actions.

The rest of this paper is as follows. Section 1 defines the problem statement. Section 2 presents the Concurrent Partial Forward Decomposition procedure (CPFD). Section 3 introduces the concrete STRIPS encoding of CPFD, called Concurrent Task Holders Decomposition encoding (CTHD). In the last section, we compare CTHD with HTN2STRIPS \cite{Alford2016BoundTP}, which is the current state-of-the-art encoding from HTN to STRIPS.

\section{Problem statement}

\subsection{STRIPS Planning Problems}

A {\it STRIPS planning problem} is a tuple $P = (L, A, I, G)$ where $L$ is a finite set of logical propositions, $A$ is a finite set of actions, $I \subseteq L$ is the initial state, and $G \subseteq L$ is the goal.

An {\it action} $a$ is a triplet $a = (pre(a), add(a), del(a))$ where $pre(a)$ is the action's {\it preconditions}, $add(a)$ is its positive {\it effects} and $del(a)$ its negative ones, each a set of propositions. Two actions $(a, b)$ are {\it independent} iff $del(a) \cap (pre(b) \cup add(b)) = \emptyset$ and $del(b) \cap (pre(a) \cup add(a)) = \emptyset$. Note that action independence only depends on action definitions. In the following, for all $a \in A$, $nInd(a)$ will denote the set of actions $b \in A$ \textit{dependent} with $a$.

A {\it state} $s$ is a set of logical propositions. The result of applying an action $a$ to state $s$ is a state $s'$ defined by the transition function $s' = \gamma (s, a) = (s - del(a)) \cup add(a)$ if $pre(a) \subseteq s$, and undefined otherwise. Let an {\it action layer} $\pi$  be a set of pairwise independent actions, and $pre(\pi) = \bigcup_{a \in \pi} pre(a)$. $add(\pi)$ and $del(\pi)$ are defined in the same way. By extension $s' = \gamma (s, \pi) = (s - del(\pi)) \cup add(\pi)$ if $pre(\pi) \subseteq s$, and undefined otherwise. Note that the actions of $\pi$ can be executed concurrently or in any sequential permutation and still yield exactly the same state $s'$.

\begin{figure}[t]
    \centering
    \includegraphics[scale=0.5]{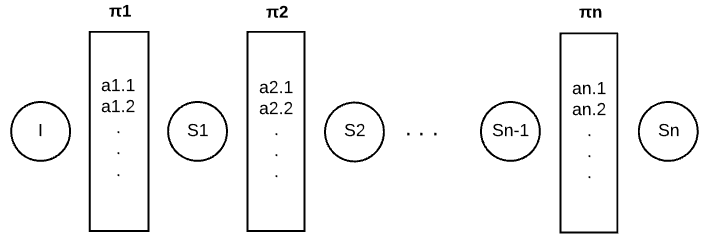}
    \caption{A layered plan with the successive states resulting from the action layer application. Circles represent states, and rectangles action layers.}
    \label{layeredPlan}
\end{figure}

A {\it layered plan} $\Pi$ is a sequence of action layers $\langle \pi_1, \ldots, \pi_n \rangle$. Let $\gamma(s, \Pi) = \gamma(\gamma(s, \pi_1), \langle \pi_2, \ldots, \pi_n \rangle)$. $\pi_i$ {\it precedes} $\pi_j$ if $i < j$. Likewise $a_i \prec a_j$ if $a_i \in \pi_i$, $a_j \in \pi_j$, and $\pi_i$ {\it precedes} $\pi_j$.
A layered plan $\Pi$ is a solution to a STRIPS planning problem $P = (L, A, I, G)$ iff $G \subseteq \gamma(s, \Pi)$ (see Fig.~\ref{layeredPlan}).

In the following, a conditional action will be used, we used the semantic defined by the ADL formalism \cite{adl}. As a regular action, a conditional one is defined by a set of preconditions. Its effects however depend on a set of conditions. For each condition verified, the corresponding effect is applied. This action is used for simplicity reasons but can easily be converted into a set of non condition actions.

\subsection{HTN Planning Problems}

We build on STRIPS planning problem definition to define a {\it HTN planning problem} as a tuple $P = (L, {\cal T}, A, M, I, tn)$ where $L$ is a finite set of logical propositions, $\cal T$ is a finite set of {\it tasks}, $A$ is a finite set of actions, $M$ is a finite set of methods, $I \subseteq L$ is the initial state and $tn$ the initial task network. There are two kind of tasks: {\it primitive} tasks that can be resolved by a STRIPS action $a = (task(a), pre(a), add(a), del(a)) \in A$ , and {\it compound} tasks, which can be recursively decomposed into either primitive or compound tasks by a method $m \in M$.

A {\it task network} is a tuple $tn = (T, \prec, \alpha)$ such that $T$ is a finite set of tasks symbols, $\alpha :T \mapsto \cal T$ maps indexes to tasks in $\cal T$, and $\prec$ is a partial order over $T$ representing precedence constraints: $t$ {\it precedes} $t'$ if $t \prec t'$, or equivalently $(t, t') \in \prec$. $\prec$ is transitive. A task $\alpha(t), t \in T$ is {\it trailing} if $\forall t' \in T$, $(t', t) \notin \prec$ ($t$ has no predecessor in $T$). $trail(tn)$ will denote the set of trailing tasks in $T$. Symmetrically, a task $\alpha(t), t \in T$ is a \textit{last task} if $\forall t' \in T, (t, t') \notin \prec$ ($t$ has no successor in $T$).

A {\it method} is a tuple $m = (task(m), pre(m), tn(m))$ where $task(m)$ is the compound task  {\it decomposed by the method} $m$, $pre(m)$ is the method's {\it preconditions} and $tn(m)$ is a task network. A method $m$ is a {\it resolver} of a compound task $\tau$ if $task(m) = \tau$. Note that a given compound task can have various methods to resolve it: $task(m) = task(m') = \dots = \tau$.

An action $a = (task(a), pre(a), add(a), del(a))$ can be applied to resolve a primitive task $\alpha(t)$ of the initial task network $tn$ if $t$ is trailing, $task(a) = \alpha(t)$ and $pre(a) \subseteq I$. The result is a new problem $P' = (L, {\cal T}, M, I', tn')$ where $I' = \gamma(I, a)$ and $tn' = (T \ \backslash \ \{t\}, \{(t', t'') \in \ \prec \ | \ t' \backslash = t\}, \alpha \backslash \{(t, \alpha(t))\})$. In a symmetrical manner, a method $m = (task(m), pre(m), tn(m))$ can be applied to resolve a compound task $\alpha(t)$ of the task network $tn$ if $t$ is trailing, $task(m) = \alpha(t)$ and $pre(m) \subseteq I$. The result of applying the method $m$ with $tn(m) = (T_m, \prec_m, \alpha_m)$ is a new problem $P' = (L, {\cal T}, M, I, tn')$ where $tn' = (T', \prec', \alpha')$ and:
\begin{eqnarray*}
    T' & = & (T \ \backslash \ \{t\}) \cup T_m \\
    \prec' & = & \{(t', t'') \in \ \prec \ | \ t'' \not = t \}\ \cup \prec_m \cup\\
           &   & \{(t'', t') \in T_m \times T \ | \ (t, t') \in \prec\}\\
    \alpha' & = & \{(t', \alpha(t')), t' \in T \backslash \{t\}\} \cup \alpha_m
\end{eqnarray*}
In other words, in $\prec'$ we keep all the precedence constraints of $\prec$ that does not involve $t$, add all the precedence constraints in $\prec_m$, and propagate precedence transitivity between $\prec$ and $\prec_m$ through $t$.

Applying either an action $a$ or a method $m$ to resolve a task in a planning problem $P$ is called a \textit{progression}. If $t_p \in \cal T$ is a primitive task of $P$, resolving $t_p$ by $a$ is a \textit{progression} denoted $P \mapsto_{t_p}^{a} P'$. Similarly, if $t_c$ is a compound task of $P$, decomposing $t_p$ using $m$ is a \textit{progression} denoted $P \mapsto_{t_c}^{m} P'$.

To conclude, a layered plan $\Pi = \langle \pi_1, \ldots, \pi_n \rangle$ is a solution for a HTN planning problem $P = (L, {\cal T}, A, M, I, tn)$ if (1) there exists a sequence of progressions that transforms $P$ into $P' = (L, {\cal T}, M, I', (\emptyset, \prec'))$ (i.e. all the tasks of $P$ have been resolved), and (2) $a_i \prec a_j \Leftrightarrow task(a_i) \prec' task(a_j)$ (i.e. action precedence constraints in the layered plan $\Pi$ are equivalent to the primitive task precedence constraints in $P'$). In section 2, we show how CPFD builds a layered solution plan by applying progressions on $P$.

\subsection{HTN to STRIPS Encoding Problems}

The Hierarchical Task Network (HTN) formalism has been shown to be more expressive than STRIPS \cite{complexityHTN}. This means that any STRIPS problem can be formulated as a HTN problem but not the other way around. Therefore, the translation of a HTN problem into a STRIPS problem is not always possible. However, it has been proven by \cite{Alford2016BoundTP} that this translation is possible if the size of the solution task network can be bounded. Given a HTN problem and a sequential solution plan, the minimum (respectively maximum) bound is the smallest (respectively largest) number of tasks in any task network visited by the sequence of progressions carried out to find this solution plan.

In practice, not all problems have a maximum bound, but all solvable problems have a minimum bound. These bounds are not directly related to the length of a problem solutions, though the minimum progression bound is smaller than the optimal plan length\footnote{For more details about the method to compute the progression bound of the solution task network see \cite{Alford2016BoundTP}}.

Our encoding also assumes the bound existence. In addition, we make two other assumptions on the HTN problem to encode:
\begin{enumerate}
    \item in the initial HTN problem, $T$ is singleton. Otherwise, it is always possible to add a root dummy-task and a dummy-method to decompose it.
    \item every method of the HTN problem has a task network with a \textit{single last task}. If a method does not have it, a dummy-task with no successor is added to the task network.
    \item methods have no preconditions. Otherwise, a trailing dummy action is added to the method task network with no effects and the method's preconditions.
\end{enumerate}

These assumptions are made without loss of generality and will simplify the notations in the following. These assumptions were also made by the current state of the art encoding HTN2STRIPS \cite{Alford2016BoundTP}.

\section{Concurrent Partial-order Forward Decomposition}

In this section, we propose a recursive procedure to solve HTN problems called CPFD (Concurrent Partial-order Forward Decomposition) and generate layered plans.

CPFD procedure is detailed in Alg. \ref{algoConc}. CPDF is an adaptation of PFD (Partial-order Forward Decomposition) \cite{ghallab} procedure to output layered plans (see Figure~\ref{layeredPlan}). A layered plan is solution of a HTN problem if actions resolve all the tasks of the task network, and if the ordering constraints of the actions in the layered plan satisfy the precedence constraints in this task network. CPFD tries to solve recursively the trailing tasks as in the PFD procedure. The difference lies on the resolution of the primitive tasks: while PFD adds actions to a sequential plan, CPFD adds them to layers of independent actions.

More precisely, CPFD takes as input four parameters: a HTN problem $P = (L, {\cal T}, A, M, I, (T,\prec))$, a layered plan, the index $i$ of the current layer $\pi_i$ and $\tau$ the set of primitive tasks resolved by the actions in $\pi_i$. The initial call of CPFD is CPFD($P, \Pi = [[]], i = 0, \tau = \emptyset$). At each recursive call, CPFD checks if the list of tasks $T$ of the task network is empty, i.e., no more tasks have to be resolved. If this condition is satisfied, the layered plan $\Pi$ is a solution to $P$ and $\Pi$ is returned. Otherwise, a task $t \in T$ is non deterministically selected among the trailing tasks (tasks without predecessors with respect to precedence constraints), and a resolver is non deterministically chosen. As in the PFD procedure, there are two ways to resolve $t$ depending on whether $t$ is primitive or compound.

\begin{description}
    \item[Case 1. (Primitive task)] The resolvers of $t$ are actions $a$ whose preconditions are satisfied in the current state $I$ and that are independent of all the actions already planned in the current layer $\pi_i$. Two cases are possible: either $t$ has no resolvers and the current layer $\pi_i$ is empty, meaning no action can solve $t$ in the current state $I$, and CPFD returns {\sc Failure}. Or $t$ has a resolver but this resolver is not an independent action in the current layer: then CPFD moves to the next layer by applying to the current state all the actions already committed in the current layer. The idea is that $t$ could be resolved by an action in a next state concurrently with other independent actions. Obviously, if $t$ has an independent resolver $a$, $a$ is added to the current layer $\pi_i$ and $t$ is added to the set of resolved primitive tasks $\tau$.
    \item[Case 2. (Compound task)] CPFD computes all the methods resolving the compound task $t$, i.e., whose preconditions are satisfied in the current state $I$. If there is no method, then $t$ cannot be solved, and CPFD returns {\sc Failure}. Otherwise, CPFD non deterministically chooses a method $m$ decomposing $t$, update the task set and the ordering constraints accordingly.
\end{description}
CPFD($P, \Pi, i, \tau$) is then called recursively until the tasks to solve in $P$ are emptied ($T = \emptyset$, line 2) or a failure condition is met (line 9 and 24).

\begin{algorithm}[!t]
\SetAlgoLined
\caption{{\sf CPFD}($P, \Pi, i, \tau$)}
\label{algoConc}
\DontPrintSemicolon
\SetKwFunction{CPFD}{\sf CPFD}

\{$P = (L,{\cal T},A, M,I,(T, \prec, \alpha))$ is the current problem\} \;
\lIf{$T = \emptyset$}{ $\textbf{return } \Pi$ }
$toSolve \gets trail(tn)\ \backslash\ \tau $\;
$\pi_i \gets get(\Pi, i) $\;
$\textit{nondeterministically choose } t \in toSolve$\;
\eIf{$\textit{t is primitive}$}{
    $resolvers \gets \{a \in A\ |\ task(a) = t, pre(a) \subseteq I\  and\ (\forall b \in \pi_i, $\text{ a independent of b})$ \}$\;
    \eIf{$resolvers = \emptyset$}{
    \eIf{$\pi_i = \emptyset$}{ \textbf{return }Failure}{
        $I \gets \gamma(I, \pi_i)$ \tcp*{Apply the layer effects}\;
        $\Pi \gets \Pi + [] $\tcp*{Add a new empty layer}\;
        $i \gets i + 1 $ \tcp*{Index of the new empty layer}\;
        $T \gets T\ \backslash\ \tau $ \tcp{Update the task network}\;
        $\prec' = \{(t', t'') \in \ \prec \ | \ t'' \not = t \}\ \cup \prec_m \cup
           \hspace{0.65cm} \{(t'', t') \in T_m \times T \ | \ (t, t') \in \prec\}$\;
        $\tau \gets \emptyset$ \tcp*{Reset the resolved tasks set}\;
        }}
        {$\textit{nondeterministically choose a} \in resolvers$\;
        $\pi_i \gets \pi_i \cup \{a\}$ \tcp*{Add $a$ to the current layer}\;
        $\tau \gets \tau \cup \{t\}$ \tcp*{Add $t$ to the resolved tasks set}\;
    	}}
    	{$\{t \textit{ is compound}\}$\;
    $resolvers \gets \{m \in M\ |\ task(m) = t\}$\;
    \lIf{$resolvers = \emptyset$}{
        \textbf{return } Failure\;
    }
    $\textit{nondeterministically choose } m \in resolvers$\;
    \{$m = (T_m,\prec_m)$\} \;
    $T \gets (T\ \backslash\ \{t\}) \cup T_m$ \tcp*{Decomposing $tn$ with $m$}\;
    $\prec \gets \{(t', t'') \in \ \prec \ | \ t'' \not = t \}\ \cup \prec_m \cup$ \;
    \hspace{0.65cm}$\{(t'', t') \in T_m \times T \ | \ (t, t') \in \prec\}$\;
}
$\textbf{return } $\CPFD{$P, \Pi, i, \tau$}\;

\end{algorithm}

\begin{theorem}{}
Concurrent HTN \textit{is sound and complete}.
\end{theorem}

\paragraph{\it Proof sketch (Soundness)} All produced plans come from a progression of the initial task network, thus there is a sequence of task decomposition that produced the primitive tasks is the solution plan. Furthermore, since a primitive task can be added to a layer if the corresponding node is unconstrained, all tasks preceding the one added have been planned on previous layers. Thus the ordering constraints in $\prec$ are satisfied in the solution plan. Thus output plans are sounds.

\paragraph{\it Proof sketch (Completeness)} We will show that CPFD is complete based on the demonstration that PFD is complete.
Let $P = (L, {\cal T}, A, M, I, tn)$ be a HTN problem and $\Pi = \langle \pi_1, \ldots, \pi_n \rangle$ a layered solution plan of $P$. Let us show that there is a sequence of recursive calls of CPFD outputting $\Pi$.
First, let us note that given a concurrent layer $\pi = \{a_1, \dots, a_k\}$, any linearization of that layer ($\langle a_{\gamma(1)}, \dots, a_{\gamma(k)}\rangle$ where $\gamma$ is a permutation function of $\{1, \dots, k\}$) is a sequence of actions which can be applied to the same states as $\pi$. This is due to the \textit{mutual independence} property of the actions within a concurrent layer.
From there, any sequential plan produced by linearizing every layer of $\Pi$ (by taking any permutation function on the layers) is a sound plan that also solves $P$. Let us consider the linearization $\Pi_l$ defined by the $n$ permutation functions $\gamma_1, \dots, \gamma_n$.
Since PFD is a complete algorithm, there is a sequence of recursion of PFD which outputs $\Pi_l$. We will show that there is an analogous sequence of CPFD recursions outputting $\Pi$.
At each recursion, PFD and CPFD either solve an unconstrained abstract task, or an unconstrained primitive task. While they solve abstract tasks the same way, PFD solves a primitive task by a adding an action resolving it to the head of the plan, meanwhile CPFD adds the action resolving the task to the concurrent layer at the head of the plan. If the task can not be resolved, PFD returns a Failure while CPFD tries to add a new concurrent layer to the plan.
Thus, for each recursive PFD call resolving an abstract task, the analogous call of CPFD is to solve the same abstract task. Each recursive call of PFD resolving a primitive task is analogous to a CPFD call adding the action to the current concurrent layer. However, CPFD requires extra recursive calls compared to PFD: it needs to select and try to resolve an unsolvable task after each layer.
In conclusion, the analogous sequence of recursion of CPFD is the one solving the same abstract and primitive tasks than PFD but with the insertion of recursive calls trying to solve an unsovlable task after resolving the last action of a layer of $\Pi$. Thus there is a sequence of CPFD recursions outputting $\Pi$ and CPFD is complete.

\subsection{Example of CPFD application}

Let us consider a simple example of application of the CPFD algorithm. Let us consider two propositions $p_1$ and $p_2$, an initial state $I = \{p_2\}$ and $tn_0 = (T_0, \emptyset, \emptyset)$ the initial task network with a single compound task $T_0$.

$T_0$ can be decomposed by a single method $m_0$ into three unordered primitive tasks $t_1$, $t_2$, and $t_3$. The task $t_1$ can be resolved by an action $a(t_1) = (t_1, \emptyset, \{p_1\}, \emptyset)$, then $a(t_2) = (t_2, \{p_1,p_2\}, \emptyset, \emptyset)$ and  $a(t_3) = (t_3, \emptyset, \emptyset, \{f_2\})$. The only solution plan is the sequential plan $\langle a(t_1), a(t_2), a(t_3)\rangle$.

When solving this problem, CPFD would first non deterministically choose a trailing task among the initial task network. There is only one, $T_0$. Then a resolver is chosen, there is only one $m_0$ which would be applied to result in a new (unordered) task network $tn = (\{t_1, t_2, t_3\}, \emptyset, \emptyset)$. On the next iteration, three tasks can be non deterministically chosen. Choosing $t_2$ or $t_2$ would lead to \textit{Failure }since they have no valid resolver in the current state and the current layer is empty. Choosing $t_1$ would offer a single valid resolver $a(t_1)$ which would be added to the current layer. On the next iteration, either $t_2$ or $t_3$ can be chosen, and choosing either one would lead to no resolver valid in the current state. However this time the current layer in not empty, so instead of returning a \textit{Failure}, CPFD would switch to the next layer by applying the effects of $a(t_1)$. On the next iteration, the updated state offers a valid resolver for $t_2$, which can be added to the next layer and so on... In the end, we obtain the sequential plan $\langle a(t_1), a(t_2), a(t_3)\rangle$

Now let us consider the same example but with $a(t_3) = (t_3, p_2, \emptyset, \emptyset)$. In that case, after decomposing $T_0$, both $t_1$ and $t_3$ have a valid resolver in the current state. These resolvers are independent, so CPFD can choose to resolve $t_1$, then $t_3$ (or $t_3$ then $t_1$) in the first layer. Then the only remaining task would be $t_2$ which have no resolver in the current state. So CPFD switches to the next layer before adding it to the second layer. In that case we produced a concurrent plan $\langle \{a(t_1), a(t_3)\}, \{a(t_2)\}\rangle$.

\section{Planning the planning: Translating HTN to STRIPS}

We present in this section the concrete STRIPS encoding of the compound algorithm previously presented, called Concurrent Task Holders Decomposition encoding (CTHD).

\subsection{Taskholder Encoding}

The CPFD procedure described previously resolves recursively unconstrained primitive tasks and compound tasks by modifying the initial task network of the problem until the task network contains only an empty set of tasks. To encode this process, we need first to model a task network with STRIPS. To achieve this, we use the concept of \textit{taskholder} introduced first by  \cite{Alford2016BoundTP}. A taskholder is a STRIPS object that will act as a container for a task. By way of extension, a task network is modeled as a stack of taskholders: static ordering relationship between taskholders defines in which order the taskholders can be allocated to the tasks during the CPFD procedure and fluent relationship define the ordering constraints between the tasks of the task network. The number of taskholders should be fixed before translating the HTN problem. Similarly to HTN2STRIPS, CTHD requires at least as many taskholders as there are tasks in the largest explored task network, thus the number of taskholders in CTHD can be estimated the same way as HTN2STRIPS estimates the number of taskholders.

In addition, we order the taskholders into a stack which defines the order in which the taskholders can be used.

The current layer of the \textit{layered plan} under construction is represented by a set of propositions, each proposition denoting the fact that an action $a \in A$ is planned in the current layer or not.

\begin{figure}[!t]
    \centering
    \includegraphics[scale=0.35]{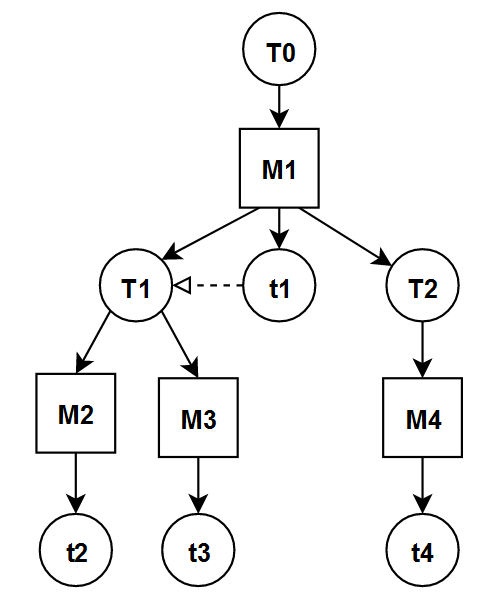}
    \caption{The task decomposition graph of $T0$. In this graph, each task node (represented with circles) is linked to the method resolving it (represented with squares). For instance, $T1$ can be decomposed into $t2$ by applying $M2$ or into $t3$ by applying $M3$. The ordering constraints are represented with the dotted arrows, so when decomposing $T0$ with $M1$, three subtasks are generated, $T1, t1$ and $T2$ where $t1$ must be planned before $T1$.}
    \label{tdgExample}
\end{figure}

\begin{figure}[!t]
    \centering
    \includegraphics[scale=0.45]{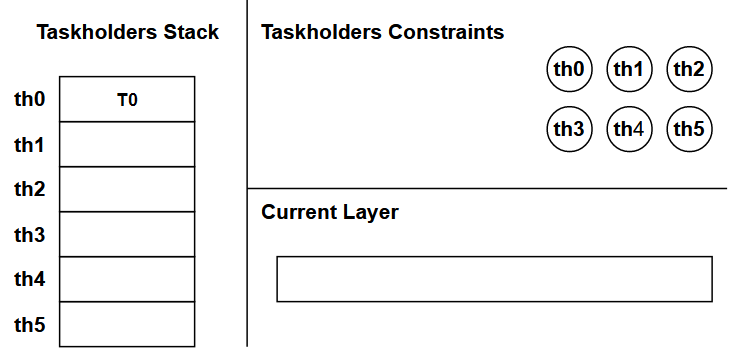}
    \caption{The problem is initialized by setting the initial task into the first taskholder.}
    \label{initialState}
\end{figure}

To fix ideas, let us consider the planning problem defined as a task decomposition graph (see Figure \ref{tdgExample}), a graphical representation of the initial structures of this problem, i.e., taskholders and current layer, is given Figure \ref{initialState}.

\subsection{Encoding as Actions}

\begin{figure}[!t]
    \centering
    \includegraphics[scale=0.45]{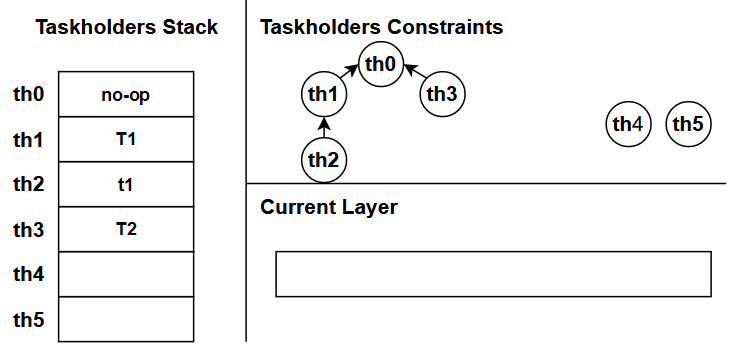}
    \caption{\textit{T0} is decomposed into four tasks, two compound ones \textit{T1} and \textit{T2}, and two primitive ones \textit{t1} an \textit{no-op}. Since neither \textit{T1}, \textit{T2} or \textit{t1} is a last task, a \textit{no-op} action is inserted instead of \textit{T0}. The constraint over the taskholders are represented on the top right graph: each directed edge represents a precedence constraint. So for instance, the task in $th2$ should planned before the one in $th1$.}
    \label{decompositionActionv3}
\end{figure}

\begin{figure}[!t]
    \centering
    \includegraphics[scale=0.45]{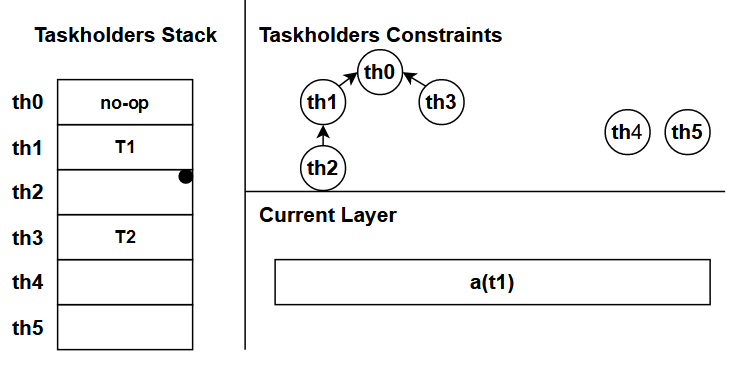}
    \caption{The second taskholder was unconstrained, and contains a primitive task. It is added to the plan step by removing the task from the taskholder, adding the action $a(t1)$ resolving $t1$ to the plan step, and marking the taskholder as resolved (represented by the black dot).}
    \label{addToStepv3}
\end{figure}

\begin{figure}[!t]
    \centering
    \includegraphics[scale=0.45]{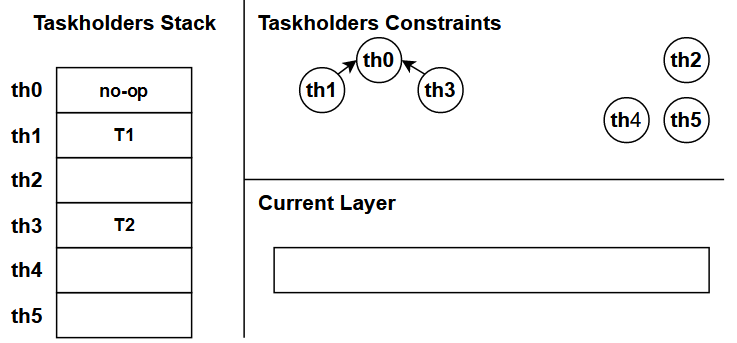}
    \caption{The plan step is terminated by emptying it, the constraints implied by the resolved taskholders are removed.}
    \label{nextStep}
\end{figure}

The dynamics of the CPFD procedure (Alg . \ref{algoConc}) is defined by three types of STRIPS actions: (1) the first type of actions resolves an unconstrained compound task and update the current task network according to a method decomposition (2) the second type of actions resolves a primitive task and add an action to the current layer;  and (3) the last type of actions is the one switching layer, making the algorithm build the next layer of the solution plan. The planning process will choose one action among the three types. Applying one of this action is equivalent to one recursive call of CPFD procedure. The planning process ends when there is no tasks left in the task network, i.e., when all taskholders are empty. These actions are defined as follows:

\begin{enumerate}

    \item \textbf{Actions for resolving compound tasks:} These actions reflect the decomposition of a compound task into subtasks according to a method. It corresponds to the lines 27 to 30 of CPFD procedure. In order to apply these actions, the following must be verified:
    \begin{itemize}
        \item The taskholder containing the decomposed task is unconstrained.
        \item There is enough taskholders remaining in the stack.
    \end{itemize}
    It works on our example as represented  Figure~\ref{decompositionActionv3}. The substasks of the method $M_1$ decomposing $T_0$ are added to the stack of taskholders and the task $T_0$ is replaced by the last task of the current task network. As $M_1$ as no \textit{last task}, the no-op action is used. Finally, the \textit{ordering} constraints are set over the taskholders.

    \item \textbf{Actions for resolving primitive tasks:} These actions resolve an unconstrained primitive task of the task network and add the resolver action $a \in A$ into the current layer (lines 21 to 23 of Alg. \ref{algoConc}).  In CPFD procedure, it translates to adding the action resolving a primitive task contained in an unconstrained taskholder to the current layer, removing the primitive task from the taskholder and marking the taskholder as resolved. These actions can be executed when the following conditions are verified:
    \begin{itemize}
        \item The preconditions of the action are satisfied.
        \item The taskholder containing the task is \textit{unconstrained}.
        \item All actions already planned at the plan step are \textit{independent} with the task.
    \end{itemize}
     An example of application is displayed on Figure \ref{addToStepv3}. The taskholder $th2$ is unconstrained and contains a primitive task $t1$ that can be resolved by the action $a(t1)$. The task is resolved by adding $a(t1)$ to the current layer, $th2$ is emptied and marked as resolved.

    As in Alg. \ref{algoConc}, the constraints implied by the resolved taskholder are not removed yet, they will when going to the next layer. 

    \item \textbf{Action for switching of plan layer:} This CTHD action corresponds to switching to the next layer. This action empty the \textit{current layer}, setting up the next one in the solution plan. It also removes the constraint implying the resolved taskholders. This action can be applied in any situation and works as displayed on Figure~ \ref{nextStep}. In this example, only $th2$ is marked as resolved. After the application of the action, the constraints implying $th2$ are removed, so the constraint between $th2$ and $th1$ is deleted, additionally $th2$ is unmarked as resolved and set as empty. Then the current layer is emptied by removing all actions in it.
\end{enumerate}

\subsection{Concurrent Taskholder Decomposition Encoding}

In this section we present the STRIPS formulation of the translated problem. Similarly to HTN2STRIPS, our translation depends on an integer parameter denoted $b$ representing the number of taskholders (i.e. the maximum number of tasks in the task network). We first define the predicates used to encode the problem, then we define the encoding of the three types of actions presented above. We end by presenting the initial state and goal state of the translated problem.\\

Let $P = (L, {\cal T}, A, M, I, tn)$ be a HTN problem and let $b \in \mathbb{N}$ the number of \textit{taskholders}. CTHD encoding generates a STRIPS problem $CPFD(P,b) = (L \cup L', A', I \cup I', G')$.\\

The encoding generates the set propositions $L'$ based on the following predicates:
\begin{itemize}
    \item $(not\_constraint\ ?th1\ ?th2 - taskholder)$ represents the fluent constraints over the taskholders. The predicate is inverted for convenience, so when the proposition $(not\_constraint\ th1\ th2)$ is false, it means that the task contained in $th1$ must be planned before the task contained in $th2$.
    \item $(empty\ ?th - taskholder)$ represents the fact that a taskholder is empty. The proposition $(empty\ th)$ is true if the taskholder $th$ does not contain a task.
    \item $(in \ ?t - \ task \ ?th\ -\ taskholder)$ is the predicate representing whether or not the task $?t \in {\cal{T}}$ is set in the parameter taskholder. The proposition $(in \ t\ th)$ is true if $t$ is set in $th$.
    \item $(not\_planned \ ?a - action)$ is true if $a$ is not planned in the current plan step.
    \item $(prec\_th\ ?th1\ ?th2 - taskholder)$ is a predicate defining the static relationship between the taskholders and defines the taskholders \textit{stack}. So the proposition $(prec\_th\ th1\ th2)$ is true is $th1$ is \textit{above} $th2$ in the stack. In the following, this order will be fixed and $\forall 0 \leq i,j < b$ the proposition $(prec\_th\ th_i\ th_j)$ will be true if and only if $i \leq j$.
    \item $(resolved\ ?th\ -\ taskholder)$ is a predicate representing either or not a taskholder have been resolved. So the proposition $(resolved\ th)$ is true if the taskholder $th$ have been resolved.\\
\end{itemize}

The encoding generates the set of actions $A' = A_{c} \cup A_{p} \cup A_{l}$ where $A_c$ is the set of actions for resolving compound tasks, $A_p$ the set of actions for resolving primitive tasks and $A_l$ the set of actions for switching of plan layer:
\begin{itemize}
    \item {\bf Actions for resolving compound tasks:} \\
    Let $m = (task(m), pre(m), tn(m))$ and  $(task_1,$ $task_2,\ldots, task_k)$ the subtasks in $tn$. We assume that $task_k$ is the \textit{last task} of $tn$. For all methods $m \in M$ there is an action $a_m \in A_c$ with $k$ parameters $(?th_1, ?th_2, \ldots, ?th_k)$ defined as follows:
    \begin{itemize}
        \item $pre(a_m) = (in \ task(m) \ ?th_1) \wedge \\
            \bigwedge_{i = 0}^{b - 1} (not\_constraint\ th_i\ ?th_1) \wedge \\
            \bigwedge_{i = 2}^{k} (prec\_th\ ?th_i\ ?th_{i + 1}) \wedge \\
            \bigwedge_{i = 2}^{k} (empty\ ?th_i)$
        \item $add(a_m) = \bigwedge_{i =  2}^{k} (in \ task_i\ ?th_i) \wedge  (in \ task_k\ ?th_1)$
        \item $del(a_m) = \bigwedge_{i = 2}^{k} (empty\ ?th_i) \wedge \\
            \bigwedge_{task_i \prec task_j} (not\_constraint\ ?th_i\ ?th_j) \wedge \\
            \bigwedge_{i = 2}^{k} (not\_constraint\ ?th_i\ ?th_0)$
    \end{itemize}
    Note that the $k-1$ last taskholder parameters are required to be \textit{ordered} according to the static relationship defined by the $prec\_th$ predicate. For instance, if three new taskholders are required to decompose the task in $th6$, $(th6\ th2\ th4\ th7)$ is a valid combination of parameters, while $(th6\ th2\ th5\ th3)$ is not.

    \item {\bf Actions for resolving primitive tasks:}\\
    For all actions $a \in A$ there is an action $a_p \in A_p$ with one parameter: a taskholder containing $p$ and denoted $?th$. The action is defined as follows:
    \begin{itemize}
        \item $pre(a_{p}) = pre(a) \wedge (in \ task(a)\ ?th) \wedge \\
            \bigwedge_{i = 0}^{b-1}(not\_constraint\ th_i\ ?th) \wedge \\
            \bigcup_{t \in nInd(p)} (not\_planned \ ?a)$
        \item $add(a_{p}) = add(a) \wedge (empty\ ?th) \wedge (resolved\ ?th)$
        \item $del(a_{p}) = del(a) \wedge (not\_planned \ ?a)$
    \end{itemize}

   \item {\bf Action for switching layer:}\\
    $A_l$ is composed of one conditional action $a_l$ with no parameter. This action has a non conditional part: emptying the current layer, and a conditional part: unconstraining and making available for a new use the resolved taskholders. It is defined as follows:
    \begin{itemize}
        \item $pre(a_{l}) = \emptyset$
        \item $add(a_{l}) = \bigwedge_{a \in A} (not\_planned \ ?a) \wedge  \\
            \forall (?th\ -\ taskholder)$ when $(resolved\ ?th), \\ \bigwedge_{i = 0}^{b - 1}(not\_constraint\ ?th\ \ th_i)$
        \item $del(a_{l}) = \emptyset$
    \end{itemize}

\end{itemize}

Finally, let us define the encoding for the initial state and goal of the translated problem. The initial state is defined by setting the initial task into the first taskholder. The remaining taskholders are set as \textit{empty} and ordered in a stack. As there is no constraint over the taskholder yet, all constraints predicate are initialized accordingly.

$I' = (in\ task_0 \ th_0)  \wedge \bigwedge_{1 \leq i < b} (empty\ th_i)\\
     \wedge \bigwedge_{1 \leq i < j < b}  (prec\_th\ th_i\ th_{j})  \wedge \\
    \bigwedge_{0 \leq i,j < b} (not\_constraint\ th_i\ th_j) \wedge \\
    \bigwedge_{a \in A} (not\_planned\ a)$

The problem is solved when all taskholders are empty, meaning that all task are planned. The goal state is defined accordingly.

$G' = \bigwedge_{i = 0}^{p-1} (empty\ th_i)$

\subsection{Differences and Similarities  with HTN2STRIPS}

If both CTHD and HTN2STRIPS encode HTN solving with taskholders, they aim at finding different kinds of solution plans. While CTHD aims at finding concurrent plans, HTN2STRIPS can only produce sequential ones. In the following, we will experimentally compare the two encodings but one has to keep in mind that CTHD produces plans with higher expressivity than the one produced by HTN2STRIPS.

 In addition, while HTN2STRIPS taskholders are unordered, CTHD orders the taskholders into a \textit{static stack} and imposes an order in which the taskholders parameters are used. The purpose of the stack is to reduce the number of valid operators in the translated problem: let us consider an example with four taskholders and a task $T_1$ set in the first taskholder $th1$. $T_1$ can be decomposed by a method $m$ into four subtasks $t_1$, $t_2$, $t_3$ and $t_4$. The HTN2STRIPS translation, translates this method into $3!$ (equivalent) actions (one for each permutation of the three newly used taskholders). Meanwhile CTHD, by constraining the taskholder a fixed order, translates this method into a single action which corresponds to the HTN2STRIPS action where all taskholders are ordered. In the general case, if $b$ is the progression bound, the number of translated actions generated from a method requiring $k$ new taskholders is equal to the number of permutations of $k$ elements among $b$ for HTN2STRIPS. While the number of generated actions is equal to the number of \textit{crescent} permutations for CTHD, which is much lower.

\section{Experimentation and Results}

\begin{figure*}[]
     \centering
     \begin{subfigure}[b]{0.45\textwidth}
         \centering
         \includegraphics[width=0.7\textwidth]{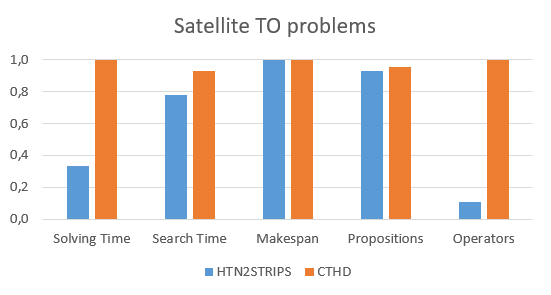}
         \label{satelliteTO}
     \end{subfigure}
     \begin{subfigure}[b]{0.45\textwidth}
         \centering
         \includegraphics[width=0.7\textwidth]{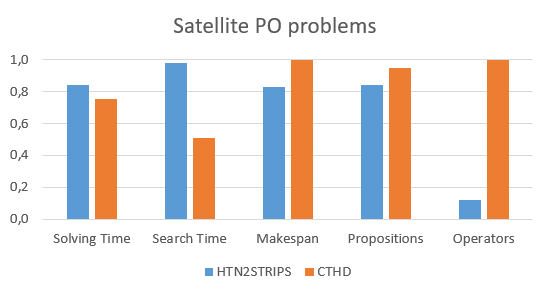}
         \label{satellitePO}
     \end{subfigure}
    \begin{subfigure}[b]{0.45\textwidth}
         \centering
         \includegraphics[width=0.7\textwidth]{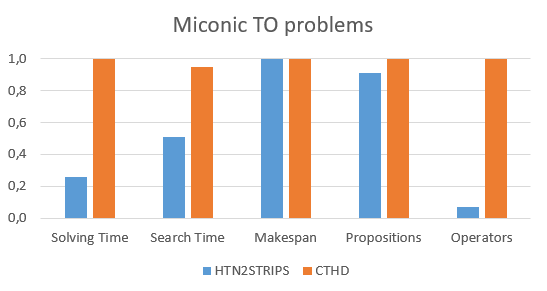}
         \label{miconicTO}
     \end{subfigure}
     \begin{subfigure}[b]{0.45\textwidth}
         \centering
         \includegraphics[width=0.7\textwidth]{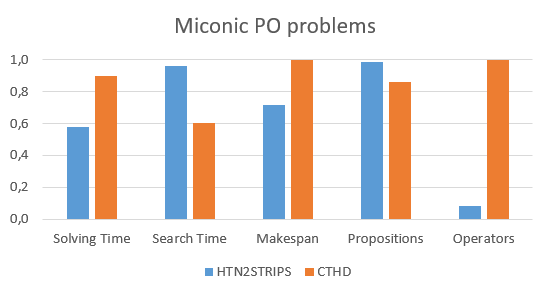}
         \label{miconicPO}
     \end{subfigure}
        \begin{subfigure}[b]{0.45\textwidth}
         \centering
         \includegraphics[width=0.7\textwidth]{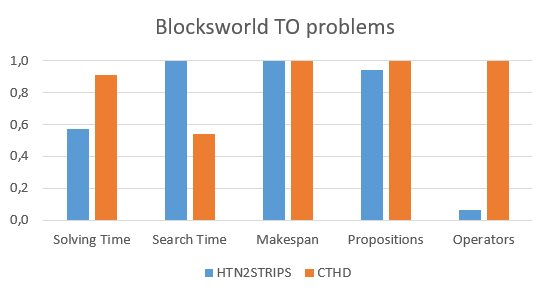}
         \label{blocksworldPO}
     \end{subfigure}
     \begin{subfigure}[b]{0.45\textwidth}
         \centering
         \includegraphics[width=0.7\textwidth]{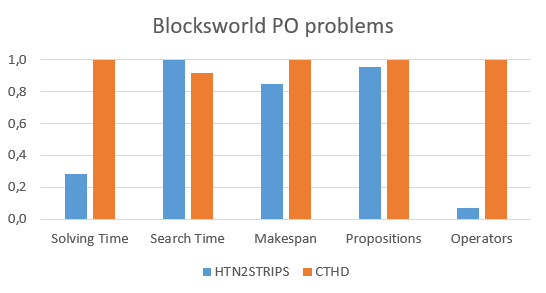}
         \label{blocksworldPO}
     \end{subfigure}
        \caption{Results for the \textit{Satellite}, \textit{Miconic}  and \textit{Blocksworld} domains: TO (Totally ordered) and PO (Partially ordered)}
        \label{satellite}
\end{figure*}

The results of the experiments are displayed on Figure \ref{satellite}.

In this section we will present the experiments and results we used to demonstrate CTHD efficiency. We compare CTHD to the current best HTN to STRIPS translation HTN2STRIPS \cite{Alford2016BoundTP}. This comparison will be made over IPC HTN benchmarks domains. These domains were chosen because they can express either totally ordered or partially ordered problems. Over each domain, the problems will be divided into totally ordered and partially ordered problems. While the solution plan of a totally ordered problem are necessarily sequential, the solution of a partially ordered problem can be concurrent.

Both HTN2STRIPS and CTHD can use the same \textit{progression bound} to solve HTN problems. So in order to have a fair comparison, all problems were solved using the same (minimal) progression bound for both encodings. Note however that if there is no benefit for HTN2STRIPS to use a progression bound bigger that the minimal one, CTHD can explore bigger task networks and thus find solution plans with higher concurrency.

The comparison of HTN2STRIPS and CTHD will be made over five metrics:

\begin{itemize}
    \item \textbf{Solving time:} This include the time spent to instantiate the problem and domain file and the time spent solving the instantiated problem.
    \item \textbf{Search time:} This corresponds solely to the time spent to solve the instantiated problem.
    \item \textbf{Solution makespan:} This corresponds to the "length" of the solution plan, meaning the number of actions within the plan for a sequential solution plan or the number of concurrent layers for a concurrent solution plan.
    \item \textbf{Number or proposition:} This corresponds to the number of proposition generated by instantiating the domain and problem file.
    \item \textbf{Number of operators:} This corresponds to the number of operators generated by instantiating the domain and problem file.
\end{itemize}

These five parameters will be evaluated by the IPC scoring metric, which is defined as follows:

$$IPC(k) = \frac{1}{|problems|} \sum_{i \in problems}\frac{min_{p \in planners}(cost(p))}{cost(k)}$$

We ran all experiments on a single core of a Intel Core i7-9850H CPU, using the Fast Downward library \cite{fastDownward} with the Delfi 1 configuration \cite{delfiIPC}, With a limit of 8GB of RAM over 600 seconds.

Over the three domains, CTHD consistently obtains a better score on the \textit{operator} metric. This is due to the difference in scaling between the number of operators in the translated HTN2STRIPS problem and the CTHD one.

\textit{Proposition} wise, both encodings are very similar. The discrepancies between domains can be explained by the number of concurrent actions within the domain: the more concurrency possible between actions, the more proposition are generated by HTN2STRIPS.

For the \textit{Makespan} metric, on totally ordered problems HTN2STRIPS and CTHD obtained the same maximal score. This is due to the fact that since both encodings use the same minimal progression bound, they both find the shortest sequential plan. On the partially ordered problems, CTHD obtains a better score than HTN2STRIPS on every domain. Since CTHD can produce concurrent plans, on partially ordered problems it is able to fit several actions into a layer, thus producing plans with smaller makespan.

When it comes to \textit{Search time} metric, the results are more mixed. Overall, HTN2STRIPS has a better \textit{Search time} for the partially ordered problems but on the totally ordered ones, HTN2STRIPS outperforms CTHD only on the \textit{Blocksworld} domain. These mixed results can be explained by the fact that a planner requires less steps to solve a HTN2STRIPS translated problem than a CTHD translated one: since HTN2STRIPS only produce sequential plans, going to the next layer is implied by resolving a primitive task. On the other hand, going to the next layer is a full solving step in a CTHD translated problem. However, we also saw that HTN2STRIPS generates more operators than CTHD, leading to higher \textit{branching factor} for the planner solving the translated problem. In the end, there is a trade off between finding a shorter plan with a higher branching factor or finding larger plans with a lower branching factor.

Finally, the \textit{Solving time} metric represents the total time spent by the planner to go from the parsing of the translated files to the solution plan. It is the sum of the time spent instantiating the files and the \textit{Search time}. On this metric, CTHD obtains a better score than HTN2STRIPS on all domains except for the partially ordered \textit{Satellite} one. However on all domains, CTHD improves its score going from \textit{Search time} to \textit{Solving time}. Once again, since CTHD requires much less operators, instantiating a CTHD translated domain file is much faster than a HTN2STRIPS one.

Overall, CTHD outperforms HTN2STRIPS both regarding the time spent to solve the translated problems and regarding the makespan of the solution plans.

\section{Conclusion and future work}

In this paper, we presented a new HTN procedure, CPFD, solving HTN problem with layered solution plan. We encoded this procedure as a STRIPS problem, thus producing a new HTN to STRIPS translation, CTHD. The translated problem, can be solved by any STRIPS planner. Then we showed experimentally that our translation outperforms the current best one HTN2STRIPS, both in terms of problem representation size, solving efficiency and quality of solution plans. This translation is a new way to solve HTN problems in a concurrent way, offering a new alternative to \textit{plan-space} HTN algorithms. However, similarly to HTN2STRIPS, our translation still depends on an integer parameter the progression bound, which bounds the size of the explored task networks. In order to improve HTN to STRIPS translation, we feel that this last point is the one to focus on.

\bibliography{biblio.bib}

\end{document}